\documentclass[conference]{IEEEtran}
\IEEEoverridecommandlockouts
\usepackage{cite}
\usepackage{amsmath,amssymb,amsfonts}
\usepackage{graphicx}
\usepackage{textcomp}
\usepackage{xcolor}
\usepackage{booktabs}
\usepackage{algorithm}
\usepackage{array}
\usepackage{algpseudocode}
\usepackage{subcaption}
\usepackage{float}
\usepackage{url}

\def\BibTeX{{\rm B\kern-.05em{\sc i\kern-.025em b}\kern-.08em
    T\kern-.1667em\lower.7ex\hbox{E}\kern-.125emX}}

\begin{document}

\title{From GNNs to Symbolic Surrogates via Kolmogorov–Arnold Networks for Delay Prediction}

\vspace{-3em}
\author{

\IEEEauthorblockN{Sami Marouani$^{1}$, Kamal Singh$^{1}$, Baptiste Jeudy$^{1}$, Amaury Habrard$^{1,2,3}$}
\IEEEauthorblockA{
\small
\textit{$^{1}$Université Jean Monnet Saint-Étienne, CNRS, Inst. d’Optique Graduate School, Lab. Hubert Curien, F-42023 Saint-Étienne, France} \\
\textit{$^{2}$Institut Universitaire de France (IUF) \  $^{3}$ Inria} \\
Email: $^{1}$\{sami.marouani, kamal.singh, baptiste.jeudy, amaury.habrard\}@univ-st-etienne.fr}
\vspace{-3em}}
\maketitle

\begin{abstract}
Accurate prediction of flow delay is essential for optimizing and managing modern communication networks.  
We investigate three levels of modeling for this task.  
First, we implement a heterogeneous GNN with attention-based message passing, establishing a strong neural baseline.  
Second, we propose FlowKANet in which Kolmogorov-Arnold Networks
replace standard MLP layers, reducing trainable parameters
while maintaining competitive predictive performance.
FlowKANet integrates KAMP-Attn (Kolmogorov--Arnold Message Passing with Attention), embedding KAN operators directly into message-passing and attention computation.
Finally, we distill the model into symbolic surrogate models using block-wise regression, producing closed-form equations that eliminate trainable weights while preserving graph-structured dependencies. The results show that KAN layers provide a favorable trade-off between efficiency and accuracy and that symbolic surrogates emphasize the potential for lightweight deployment and enhanced transparency.
\end{abstract}
\vspace{0.1cm}


\section{Introduction}

Flow delay is a key performance metric in communication networks, influencing congestion control, Traffic Engineering (TE), and Quality of Service (QoS).
Network performance prediction has traditionally relied on analytical models and Discrete Event Simulation (DES).
Queueing models make strong assumptions, while simulations face scalability and runtime limitations.  
In recent years, data-driven methods have emerged as an alternative, with Graph Neural Networks (GNNs) showing strong ability to capture dependencies between flows, links, and topologies.  

Despite their success, GNN-based models face two main challenges.  
First, they are often parameter-heavy with tens of thousands of weights, hindering deployment in resource-constrained environments.  
Second, they remain largely black-box models: while accurate, they provide little transparency about how input features combine to yield predicted delays, limiting trust and interpretability in operational use.  

Recently, Kolmogorov--Arnold Networks (KANs)~\cite{liu2024kan} have emerged as a promising alternative to conventional multi-layer perceptrons (MLPs).  
Grounded in the Kolmogorov--Arnold representation theorem, KANs replace fixed activations with learnable spline functions, enabling smooth and interpretable functional modeling.  
Their transparent structure makes them particularly suitable for analyzing relationships in network performance data.   

In this work, we investigate a spectrum of models that balance precision, compactness, and interpretability.  
Our contributions are:  
\begin{itemize}
    \item A heterogeneous GNN with attention-based message passing as a strong baseline for flow delay prediction.  

        \item \textbf{FlowKANet}, a fully KAN-based GNN architecture in which all components are implemented with spline-based operators for greater functional consistency, efficiency, and interpretability. The model incorporates \textbf{KAMP-Attn}, a mechanism that leverages KAN operators to compute both feature transformations and attention coefficients within the message-passing process.  
           
    \item A symbolic distillation of FlowKANet through block-wise regression, yielding compact analytical surrogates that preserve graph dependencies and enable lightweight, interpretable deployment.  
\end{itemize}
Our source code is available online \footnote{\url{https://github.com/marouanisami/FlowKANet-Symbolic}}.

\section{Related Works}

\subsection{Traditional Network Modeling}

Analytical queuing models, such as M/M/1 and M/M/k systems, provide closed-form expressions for metrics like average delay. While those models are mathematically elegant, they depend on strong assumptions (e.g., Poisson arrivals, exponential service times). In contrast, machine learning (ML) approaches learn directly from data, allowing more flexible and adaptive modeling of complex network behaviors.
DES approaches, on the other hand, can capture detailed protocol dynamics and queuing interactions with high fidelity. 
They are widely used in academia and industry, but their computational complexity is prohibitive: DES is inherently difficult to parallelize, scales poorly with network size, and remains unsuitable for real-time applications.
These limitations motivate the shift toward data-driven ML approaches, which learn performance models directly from traffic traces without restrictive assumptions.

\subsection{Machine Learning for Networking}

Machine learning has been used for various networking tasks such as traffic classification, anomaly detection, routing, and resource allocation ~\cite{almasan2022deep,ridwan2021applications}. However, these approaches typically treat flows as independent samples, failing to capture the inherent graph structure of communication networks. This motivates the use of GNNs, which represent networks as graphs and capture node, link, and flow dependencies. 

\subsubsection{GNN-based Models}

GNNs process graph-structured data by iteratively propagating and aggregating information between neighboring nodes through message passing~\cite{zhou2020graph,gilmer2017neural}.
In general, a GNN layer can be expressed as:
\begin{equation}
h_v^{(k+1)} = \phi^{(k)} \Big( h_v^{(k)}, \; \mathrm{AGG}_{u \in \mathcal{N}(v)} \psi^{(k)}(h_v^{(k)}, h_u^{(k)}, e_{uv}) \Big)
\end{equation}
where $e_{uv}$ are edge features, $\mathrm{AGG}$ is a permutation-invariant aggregation operator (e.g., sum, mean, max), $h_v^{(k)}$ is the feature of node $v$ at layer $k$,  $\psi^{(k)}$ is the message function,  and $\phi^{(k)}$ is the update function.  
This formulation suits communication networks, where the interaction between flows and links can naturally be represented as a graph.   

In traffic engineering (TE), GNNs have shown strong potential for resource allocation~\cite{xu2023teal, almasan2022deep}. TEAL~\cite{xu2023teal} combined a GNN with reinforcement learning and ADMM for WAN optimization, achieving near-optimal results. 
Building on this direction, FlowAttune~\cite{marouani2024advanced} applied graph attention to dynamically weight neighboring nodes during message passing.
Formally, in a Graph Attention Network (GAT)~\cite{velivckovic2017graph}, the message from a neighbor $u \in \mathcal{N}(v)$ to node $v$ is weighted by an attention coefficient
\begin{equation}
    \alpha_{vu} = \frac{\exp \big( \mathrm{LeakyReLU}(a^\top [W h_v \, \Vert \, W h_u]) \big)}{\sum_{k \in \mathcal{N}(v)} \exp \big( \mathrm{LeakyReLU}(a^\top [W h_v \, \Vert \, W h_k]) \big)},
\end{equation}
where $h_v$ and $h_u$ are node features, $W$ is a learnable weight matrix, $a$ is the attention vector, and $\Vert$ denotes concatenation.  
The updated node representation is then obtained as
\begin{equation}
    h_v' = \sigma \left( \sum_{u \in \mathcal{N}(v)} \alpha_{vu} W h_u \right),
\end{equation}
where $\sigma$ is a non-linear activation function.  

Attention mechanisms enable GNNs to capture structural dependencies and adapt to dynamic traffic, offering a flexible alternative to fixed aggregation functions. Both methods rely on modeling the network as a flow–link bipartite graph, which facilitates message passing between flows and their associated links.
This representation has inspired our own work, where we adopt a similar bipartite modeling strategy for flow delay prediction.  
GNNs have also been applied to performance prediction. 
RouteNet~\cite{rusek2020routenet,ferriol2023routenet} and its extensions have estimated end-to-end metrics such as delay and jitter with high accuracy across unseen topologies, while the GNNet Challenge~\cite{guemes2023building} further validated GNN-based performance prediction using real traffic traces, adopting RouteNet-Fermi as its baseline. These works highlight the versatility of GNNs in networking, spanning from traffic allocation to performance prediction. However, existing GNNs still face (i) large parameter counts, leading to high training and inference costs; (ii) limited scalability on large graphs; and (iii) lack of transparency, which constrains their use in operational and real-time networks.

\subsubsection{Kolmogorov--Arnold Networks (KANs)}

KANs~\cite{liu2024kan} are inspired by the Kolmogorov--Arnold representation theorem, which expresses any multivariate function as a composition of univariate ones, KANs replace fixed activations with trainable spline operators.
KAN layer computes:
\begin{equation}
    y = W \, \phi(x),
\end{equation}
where $\phi(\cdot)$ is a learnable spline function defined on a grid.
This design enables smoother function approximation, improved parameter efficiency, and greater transparency of learned transformations. 
KANs have shown promising results in scientific machine learning and physics-informed tasks~\cite{ji2024comprehensive,rigas2024adaptive,somvanshi2025survey}, but remain unexplored in networking or combined with GNNs. This motivates one of the main axes of our study: exploring the integration of KAN layers within GNN architectures for flow delay prediction.

\section{Framework for Flow Delay Prediction}

We introduce a unified framework that provides a single pipeline from raw network data to graph-based models. 
Unified here means that the same data representation, preprocessing, and normalization steps are shared across architectures, ensuring that improvements can be attributed to the model design itself. 
The framework starts with two important steps: constructing a heterogeneous bipartite graph that captures flow--link relationships, and selecting a compact set of relevant features that improves efficiency and generalization. 
On this foundation, we implement two predictive models: a baseline GNN and a KAN-augmented GNN.

\subsection{Graph Construction}

The network is represented as a heterogeneous bipartite graph 
$\mathcal{G} = (\mathcal{V}_f \cup \mathcal{V}_l, \mathcal{E})$, 
where $\mathcal{V}_f$ denotes the set of flow nodes, each corresponding to a unidirectional flow characterized by features describing its traffic profile (e.g., rate, packet size, burstiness, loss), and $\mathcal{V}_l$ denotes the set of link nodes, each representing a physical link annotated with its capacity and load. The edge set $\mathcal{E}$ connects flows to the links they traverse, as determined by routing, and includes both $(f \to l)$ and $(l \to f)$ directions to enable bidirectional message passing. This construction captures dependency of flows on the sequence of links along their path and vice versa. 

\paragraph*{Data extraction and feature engineering}
The raw dataset provides per-flow, per-link, and topology-level information. 
From this, we build feature vectors for each node type.
Flow nodes include basic attributes such as average traffic, number of packets, mean packet size, flow type, and path length, augmented with distributional descriptors that capture burstiness and variability, including inter-packet gap (IPG) statistics (mean, variance, and selected percentiles), packet-size percentiles, packet loss ratio, variance of packet sizes, inter-burst gap (IBG), rate, per-burst bitrate, and type of service (ToS). 
Each link node is annotated with its capacity and a normalized load, computed as
    \begin{equation}
        L_\ell = \frac{\sum_{f \in \mathcal{F}(\ell)} \mathrm{traffic}(f)}{C_\ell \cdot 10^9 + \varepsilon},
    \end{equation}
where $\mathcal{F}(\ell)$ is the set of flows traversing link $\ell$ and $C_\ell$ is the link capacity (Gbps). The concatenation $[\,C_\ell, L_\ell\,]$ forms the feature vector for link~$\ell$. This enriched representation incorporates not only average values but also distributional characteristics of packet timing and size for accurately modeling flow delay.

\subsection{Feature Selection}

The flow feature set exceeds one hundred dimensions, many of which are correlated or redundant. 
To reduce complexity and improve generalization, we apply Sequential Forward Selection (SFS) with a linear regression proxy and three-fold cross-validation, using mean squared error (MSE) for stable optimization on small delay values.
SFS incrementally adds features that maximize performance gain until convergence, yielding a compact subset of 16 flow features (Table~\ref{tab:selected_features}) that balance expressiveness and efficiency.

\begin{table}[h]
\centering
\caption{Selected flow features after Sequential Forward Selection.}
\label{tab:selected_features}
\begin{tabular}{ll}
\toprule
Feature name & Description \\
\midrule
\texttt{flow\_traffic} & Average flow bitrate \\
\texttt{flow\_packets} & Number of packets in the flow \\
\texttt{flow\_packet\_size} & Mean packet size \\
\texttt{flow\_type} &  CBR or MB \\
\texttt{flow\_length} & Path length (hop count) \\
\texttt{flow\_p10PktSize} & 10th percentile of packet size \\
\texttt{flow\_tos} & Type of Service (ToS) field \\
\texttt{flow\_packet\_loss} & Packet loss ratio (\%) \\
\texttt{ibg} & Inter-burst gap \\
\texttt{rate} & Flow generation rate \\
\texttt{flow\_bitrate\_per\_burst} & Average bitrate per burst \\
\texttt{flow\_ipg\_mean} & Mean inter-packet gap \\
\texttt{flow\_ipg\_var} & Variance of inter-packet gap \\
\texttt{IPG percentile P11,} & 11th, 99th, 100th percentile of  \\
\texttt{~~P99, P100} & IPG distribution \\
\bottomrule
\end{tabular}
\end{table}

\paragraph*{Feature normalization}
To stabilize training and ensure comparable scaling across heterogeneous features, we apply min–max normalization to all flow attributes:
\begin{equation}
    \tilde{x}_f = (x_f - \mathbf{m}_{\min}) \odot (\mathbf{m}_{\max} - \mathbf{m}_{\min})^{-1},
\end{equation}
where $\mathbf{m}_{\min}$ and $\mathbf{m}_{\max}$ are the per-feature minima and maxima computed on the training set. 
These statistics are stored in model buffers (\texttt{min\_feat}, \texttt{inv\_range}) and reused during inference, ensuring consistent scaling across datasets.

\subsection{Baseline GNN Model Architecture}
\begin{figure*}[!t]
    \centering
    \includegraphics[width=0.95\textwidth]{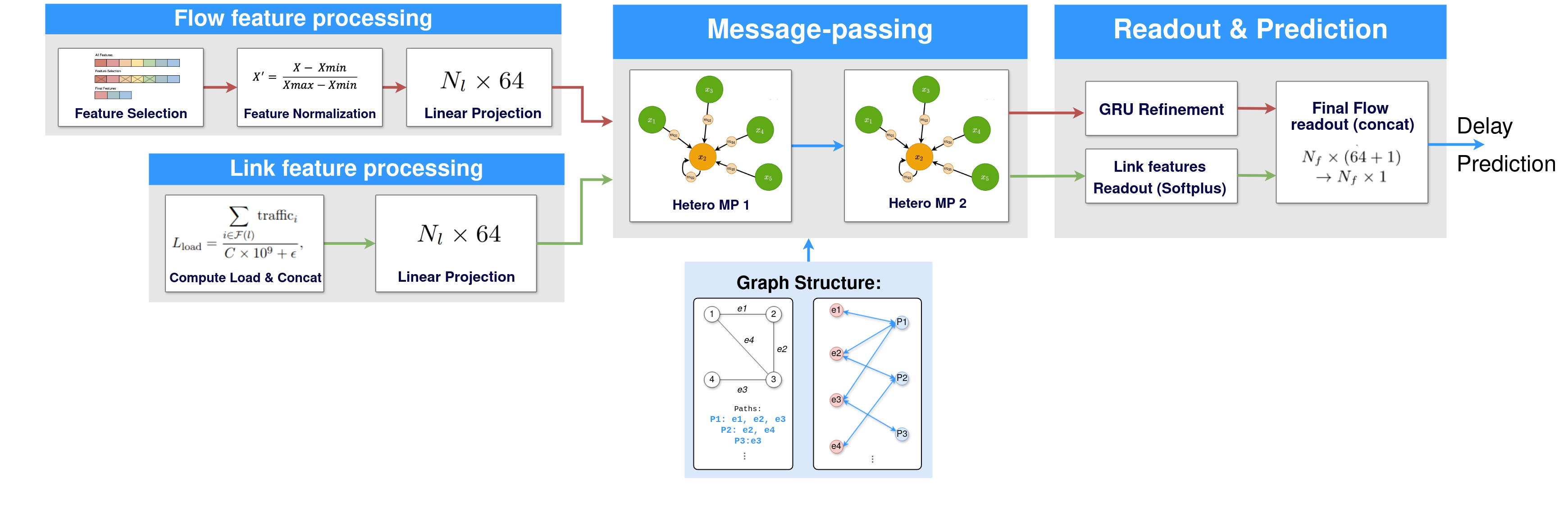}
    \caption{Baseline GNN architecture with heterogeneous message passing, attention, and GRU refinement.}
    \label{fig:gnn_architecture}
    \vspace{-1.9em}

\end{figure*}

The baseline architecture is a heterogeneous GNN designed for flow delay prediction. Its workflow is given in Algorithm~\ref{alg:gnn_forward}, while the main architectural blocks are shown in Figure~\ref{fig:gnn_architecture}.

\begin{algorithm}[h]
\small
\caption{Forward Pass of the GNN Baseline}
\label{alg:gnn_forward}
\begin{algorithmic}[1]
\State Encode flows: $h_f^{(0)} \gets \mathrm{MLP}_f(x_f)$,  and links: $h_\ell^{(0)} \gets \mathrm{MLP}_\ell(x_\ell)$
\For{$k = 1$ to $K$}
    \For{each edge $(f,\ell) \in \mathcal{E}$}
        \State Compute attention weight $\alpha_{f\ell}^{(k)}$
        \State Compute message $m_{f\ell}^{(k)} \gets \psi\!\big(h_f^{(k)},\, h_\ell^{(k)},\, \alpha_{f\ell}^{(k)}\big)$
    \EndFor
    \For{each node $v \in \mathcal{V}_f \cup \mathcal{V}_\ell$}
        \State Aggregate messages $M_v^{(k)} \gets \sum_{u \in \mathcal{N}(v)} m_{uv}^{(k)}$
        \State Update node state $h_v^{(k+1)} \gets \phi\!\big(h_v^{(k)},\, M_v^{(k)}\big)$
    \EndFor
\EndFor
\State Refine flow embeddings using GRU
\State Fuse $\big[\,h_f^{(K)} \;\Vert\; \mathrm{agg}(h_\ell^{(K)})\,\big]$
\State Predict delay $\hat{d}_f \gets \mathrm{MLP}_{\mathrm{readout}}(\cdot)$
\end{algorithmic}
\end{algorithm}

\begin{itemize}
    \item \textbf{Flow and Link Encoders:} Raw flow and link features are projected into latent embeddings of dimension $d_h$.
    
    \item \textbf{Message Passing:} Each link aggregates messages from incident flows and each flow from its traversed links.  
    Attention mechanisms compute flow-to-link and link-to-flow scores, weighting contributions by traffic intensity and congestion.
    
    \item \textbf{Readout and Prediction:} A gated recurrent unit (GRU) refines flow embeddings across iterations, followed by fusion of flow and aggregated link embeddings.  
    A Softplus layer outputs the final delay prediction.
\end{itemize}

\subsection{FlowKANet Model Architecture}

To reduce complexity while preserving expressivity, we extend the baseline by replacing all MLPs with KANs layers. The overall message passing structure remains identical, but every transformation block is spline-based. The workflow is summarized in Algorithm~\ref{alg:kan_forward}.

\begin{itemize}
    \item \textbf{Flow and Link Encoders:} Flow and link features are projected into latent embeddings by KAN layers. These initial encoders provide compact hidden representations tailored to each node type.
    \item \textbf{KAMP-Attn (Kolmogorov--Arnold Message Passing with Attention):} Messages are exchanged between flows and links using KAN operators for both transformation and attention, ensuring bidirectional propagation of information across the bipartite graph.
    \item \textbf{Readout and Prediction:} After message passing, flow embeddings are fused with their aggregated link representations and passed through a final KAN block with Softplus activation to predict the per-flow delay.
\end{itemize}

\paragraph*{KAN-based message passing}
Let $\mathbf{h}_f^{(k)}$ and $\mathbf{h}_\ell^{(k)}$ denote the embeddings of flow $f$ and link $\ell$ at iteration $k$.  
For each edge $(f,\ell) \in \mathcal{E}$, the message is computed using two \textit{shared} spline-based operators:  
(i) a \textit{transformation operator} $\mathcal{T}^{\mathrm{KAN}}_{\mathrm{f\to l}}$ that maps flow embeddings into the link space, and  
(ii) an \textit{attention operator} $\mathcal{A}^{\mathrm{KAN}}_{\mathrm{f\to l}}$ that produces edge-specific importance weights.  
These operators are shared across all edges in the same direction. 
Formally,
\begin{align}
\tilde{\mathbf{h}}_{f\ell}^{(k)} &= 
    \mathcal{T}^{\mathrm{KAN}}_{\mathrm{f\to l}}\!\big(\mathbf{h}_f^{(k)}\big), \\
s_{f\ell}^{(k)} &= 
    \mathcal{A}^{\mathrm{KAN}}_{\mathrm{f\to l}}\!\Big(
        \mathrm{LeakyReLU}\!\big(\mathbf{h}_\ell^{(k)} + \tilde{\mathbf{h}}_{f\ell}^{(k)}\big)
    \Big), \\
\alpha_{f\ell}^{(k)} &= 
    \frac{\exp(s_{f\ell}^{(k)})}{\sum_{f' \in \mathcal{N}(\ell)} \exp(s_{f'\ell}^{(k)})}.
\end{align}
The aggregated message at a node $v$ is then
\begin{equation}
    \mathbf{M}_v^{(k)} = \sum_{u \in \mathcal{N}(v)} 
    \alpha_{uv}^{(k)} \,\tilde{\mathbf{h}}_{uv}^{(k)} ,
\end{equation}
and the node embedding is updated with a residual connection:
\begin{equation}
    \mathbf{h}_v^{(k+1)} = \mathbf{h}_v^{(k)} + \mathbf{M}_v^{(k)}.
\end{equation}
This mechanism operates in both directions: flows send messages to links via $\mathcal{T}^{\mathrm{KAN}}_{\mathrm{f\to l}}$ and $\mathcal{A}^{\mathrm{KAN}}_{\mathrm{f\to l}}$, while links send messages back to flows through distinct operators $\mathcal{T}^{\mathrm{KAN}}_{\mathrm{l\to f}}$ and $\mathcal{A}^{\mathrm{KAN}}_{\mathrm{l\to f}}$. Each operator is shared across all edges of its respective direction, ensuring consistency and avoiding edge-specific parameterization.

\begin{algorithm}[h]
\small
\caption{Forward Pass of the FlowKANet}
\label{alg:kan_forward}
\begin{algorithmic}[1]
\State Encode flows and links: $\mathbf{h}_f^{(0)} \gets \mathrm{KAN}_f(x_f)$, $\mathbf{h}_\ell^{(0)} \gets \mathrm{KAN}_\ell(x_\ell)$
\For{$k = 1$ to $K$}
    \For{each edge $(f,\ell) \in \mathcal{E}$}
        \State $\tilde{\mathbf{h}}_{f\ell}^{(k)} \gets \mathcal{T}^{\mathrm{KAN}}_{\mathrm{f\to l}}(\mathbf{h}_f^{(k)})$
        \State $\alpha_{f\ell}^{(k)} \gets \mathcal{A}^{\mathrm{KAN}}_{\mathrm{f\to l}}\!\big(\mathrm{LeakyReLU}(\mathbf{h}_\ell^{(k)} + \tilde{\mathbf{h}}_{f\ell}^{(k)})\big)$
    \EndFor
    \For{each node $v \in \mathcal{V}_f \cup \mathcal{V}_\ell$}
        \State $\mathbf{M}_v^{(k)} \gets \sum_{u \in \mathcal{N}(v)} \alpha_{uv}^{(k)}\,\tilde{\mathbf{h}}_{uv}^{(k)}$
        \State $\mathbf{h}_v^{(k+1)} \gets \mathbf{h}_v^{(k)} + \mathbf{M}_v^{(k)}$
    \EndFor
\EndFor
\State $\mathbf{c}_f^{(K)} \gets \sum_{\ell \in \mathcal{N}(f)} \mathbf{h}_\ell^{(K)}$ 
\State $\mathbf{z}_f^{(K)} \gets g_{\mathrm{fuse}}\!\big[\mathbf{h}_f^{(K)} ; \mathbf{c}_f^{(K)}\big]$
\State Predict delay $\hat{d}_f \gets g_{\mathrm{final}}\!\big(\mathbf{h}_f^{(K)} + \mathbf{z}_f^{(K)}\big)$
\end{algorithmic}
\end{algorithm}

\paragraph*{Concise single-step composition}
Combining transformation, aggregation, and fusion, the per-flow output after $K$ rounds of bidirectional message passing can be expressed as
\begin{equation}
\hat{d}_f =
g_{\mathrm{final}}\!\Big(
    \mathbf{h}_f^{(K)} +
    g_{\mathrm{fuse}}\!\big[
        \mathbf{h}_f^{(K)} ; \mathbf{c}_f^{(K)}
    \big]
\Big),
\end{equation}
where $\mathbf{h}_f^{(K)}$ and $\mathbf{h}_\ell^{(K)}$ are the final flow and link embeddings, and 
$\mathbf{c}_f^{(K)} = \sum_{\ell \in \mathcal{N}(f)} \mathbf{h}_\ell^{(K)}$
is the aggregated link context. 
Here, $g_{\mathrm{fuse}}(\cdot)$ and $g_{\mathrm{final}}(\cdot)$ denote the KAN fusion and readout blocks, respectively, with the latter ending in a Softplus activation to ensure non-negative delay predictions.

\subsection{Symbolic Surrogate Models}

Although the KAN-augmented GNN is lighter than the MLP baseline, it still contains many trainable parameters.
To further reduce deployment overhead, we distill the trained FlowKANet into symbolic surrogate models, replacing each KAN block with an analytical expression that approximates its learned mapping, yielding a fully symbolic pipeline from input features to predicted delay. We employ PySR~\cite{cranmer2024pysr,cranmer2023interpretablemachinelearningscience}, combined with Optuna-based hyperparameter optimization, to discover compact analytical expressions that closely match the outputs of the corresponding KAN operators.

\paragraph{Sequential block-wise search}
The symbolic distillation is performed progressively, one block at a time, following the network structure.  
At each step, previously symbolized equations are frozen while downstream components remain neural.  
This ensures consistency of symbolic dependencies and yields a coherent chain of analytical transformations.  
The procedure can be summarized as follows:
\begin{enumerate}
    \item For each block $b$, freeze all previously symbolized blocks and keep downstream blocks neural.  
    \item Fit a PySR regressor to approximate the KAN output of $b$, while Optuna tunes PySR hyperparameters (population size, mutation rate, parsimony).  
    \item Evaluate each candidate expression inside the full model on a validation subset, fix the best formula, and proceed to the next block $b{+}1$.  
\end{enumerate}

\paragraph{Final surrogate pipeline.}
Once all blocks have been symbolized, we obtain a fully analytical model that respects the underlying graph structure.  
Formally, the surrogate prediction takes the form:
\begin{equation}
    \hat{d}_f = 
    \mathcal{E}_{\mathrm{symbolic}}\!\Bigl(
        x_f, [C_\ell, L_\ell], 
        \{\mathcal{N}(f), \mathcal{N}(\ell)\}
    \Bigr),
\end{equation}
where $\mathcal{E}_{\mathrm{symbolic}}$ denotes the composed surrogate equations, $x_f$ are the selected flow features, $[C_\ell,L_\ell]$ are link descriptors, and $\{\mathcal{N}(f), \mathcal{N}(\ell)\}$ encodes the neighborhood relations in the bipartite flow--link graph.  In other words, the symbolic surrogate maintains the same message-passing dependencies as the neural FlowKANet: flow delay predictions depend not only on local features but also on the aggregated symbolic contributions of neighboring links and flows. This yields a fully analytical yet graph-aware predictor that mirrors the inductive bias of the original architecture while eliminating the need for neural inference.
\section{Performance Evaluation}

\subsection{Experimental Setup}
We use the GNNet Challenge dataset~\cite{guemes2023building}, which provides realistic topologies and flow-level traces for graph-based performance prediction. All experiments employ the heterogeneous bipartite graph representation described in Section~III. The dataset is accessed through an API that exposes per-flow, per-link, and topology-level features, and is split into training (70\%), validation (15\%), and test (15\%) sets.

\subsection{Hyperparameter Search with Optuna}

We employ Optuna to automatically select the main architectural hyperparameters of the FlowKANet.  
The search space includes:
the hidden dimensions of flow and link embeddings, the number of message-passing layers $K$, 
KAN parameters (grid size $G$, spline order $k$, and scaling $\sigma$), as well as dropout rate, learning rate, 
and activation configuration.  

\begin{table}[h]
\centering
\caption{Best KAN parameters per block (Optuna).}
\label{tab:optuna_params_blocks}
\begin{tabular}{l>{\centering\arraybackslash}p{1.5cm}>{\centering\arraybackslash}p{1.5cm}>{\centering\arraybackslash}p{2cm}}
\toprule
Block & Grid $G$ & Order $k$ & Scale $\sigma$ \\
\midrule
\texttt{flow\_init} & 9 & 3 & 0.93 \\
\texttt{link\_init} & 7 & 5 & 1.66 \\
\midrule
$\text{flow}\!\to\!\text{link}$ (i=0) & 5 & 3 & 0.55 \\
$\text{flow}\!\to\!\text{link}$ (i=1) & 6 & 4 & 0.70 \\
$\text{flow}\!\to\!\text{link}$ (i=2) & 8 & 4 & 0.82 \\
\midrule
$\text{link}\!\to\!\text{flow}$ (i=0) & 7 & 3 & 0.73 \\
$\text{link}\!\to\!\text{flow}$ (i=1) & 7 & 5 & 0.77 \\
$\text{link}\!\to\!\text{flow}$ (i=2) & 10 & 3 & 0.33 \\
\midrule
\texttt{fuse} & 6 & 5 & 1.15 \\
\texttt{final} & 10 & 5 & 2.28 \\
\bottomrule
\end{tabular}
\end{table}

We tested several activation functions (\texttt{ReLU}, \texttt{SiLU}, \texttt{Softplus}, \texttt{Tanh}) and 
four placement strategies: \textbf{final\_only} (after the fusion block), \textbf{except\_mp} (all blocks except message passing), 
\textbf{all} (every KAN block), and \textbf{no\_activation} (none applied). Regardless of the chosen strategy, the last readout block always applies \texttt{Softplus} to guarantee non-negative flow delay predictions.  
Each Optuna trial was trained for up to 150 epochs with early stopping on the validation set.  
The Tree-structured Parzen Estimator (TPE) sampler was used for efficient exploration of the large search space.  
Table~\ref{tab:optuna_params} summarizes the best global hyperparameters, while Table~\ref{tab:optuna_params_blocks} details the KAN-specific settings for each block.  
\begin{table}[h]
\centering
\caption{Best global FlowKANet hyperparameters (Optuna).}
\label{tab:optuna_params}
\begin{tabular}{>{\centering\arraybackslash}p{2.1cm} 
                >{\centering\arraybackslash}p{1cm} 
                >{\centering\arraybackslash}p{4.5cm}}
\toprule
Parameter & Best Value & Description \\
\midrule
Flow hidden dim. & 8 & Size of flow embedding \\
Link hidden dim. & 2 & Size of link embedding \\
MP layers $K$ & 3 & Number of heterogeneous MP layers \\
Dropout & 0.1 & Regularization between layers \\
Learning rate & 0.002 & Optimizer step size \\
Activation type & \texttt{Tanh} & Activation applied to KAN outputs \\
Activation mode & \texttt{except\_mp} & Applied to all blocks except MP \\
\bottomrule
\end{tabular}
\end{table}

\subsection{Symbolic Surrogate Search}
\begin{table}[!htb]
\centering
\caption{Symbolic surrogate search space and constraints.}
\label{tab:pysr_search}
\begin{tabular}{p{3cm} p{5cm}}
\toprule
\textbf{Component} & \textbf{Options / Constraints} \\
\midrule
Binary operators & $\{+, -, \times\}$ , $\{+, -, \times, \div\}$,  $\{+, -, \times, \div, \hat{}\}$ with exponent range $[-1,2]$ \\
\midrule
Unary operators  & $\{\exp, \log, |\cdot|\}$ ,  $\{\exp, \log, \tanh, |\cdot|\}$, $\{\exp, \log, \tan, \tanh, |\cdot|\}$, \\ 
&$\{\exp, \log, \sin, \cos, \tan, \tanh, |\cdot|\}$ \\
\midrule
Expression size  & \texttt{maxsize} $\in \{7, 14, 21, 28, 35\}$ \\
\midrule
Numerical guards & $\log(\max(x,\varepsilon)),\ \varepsilon=10^{-8}$;  $\exp$ clipped to $[-50, 50]$; NaN/$\pm\infty$ replaced \\
\bottomrule
\end{tabular}
\end{table}
We apply the symbolic distillation procedure described in Section~III-E to the trained FlowKANet model.  
For each KAN block, PySR performs symbolic regression guided by Optuna-based hyperparameter optimization, 
jointly tuning the operator sets, tree complexity (\texttt{maxsize}), population size, iteration count, 
and parsimony coefficient to minimize the validation MSE.  
Expressions exceeding the desired complexity are penalized via the parsimony term, and model selection favors the most accurate symbolic representations. Numerical robustness is enforced through safe $\log/\exp$ operators and replacement of non-finite values.  
The search runs in parallel through a shared Optuna RDB and is limited to 250 trials per block.  
FlowKANet weights remain frozen during distillation.  
For each block, input–output activations are sampled from the training graphs, with a fraction $\gamma$, set to $0.5$ in our runs used for symbolic fitting and the remainder for validation within the hybrid model.  
The best expression per block is validated in context and fixed, progressively replacing all KAN modules to obtain a fully symbolic surrogate of the network.

\subsection{Predictive Accuracy}

We compare the baseline GNN, the KAN-augmented GNN, and the symbolic surrogate distilled from FlowKANet, reporting MSE and $R^2$ on the test set.

\begin{figure*}[!htbp]
\centering
\begin{subfigure}{0.30\textwidth}
\centering
\includegraphics[width=\linewidth]{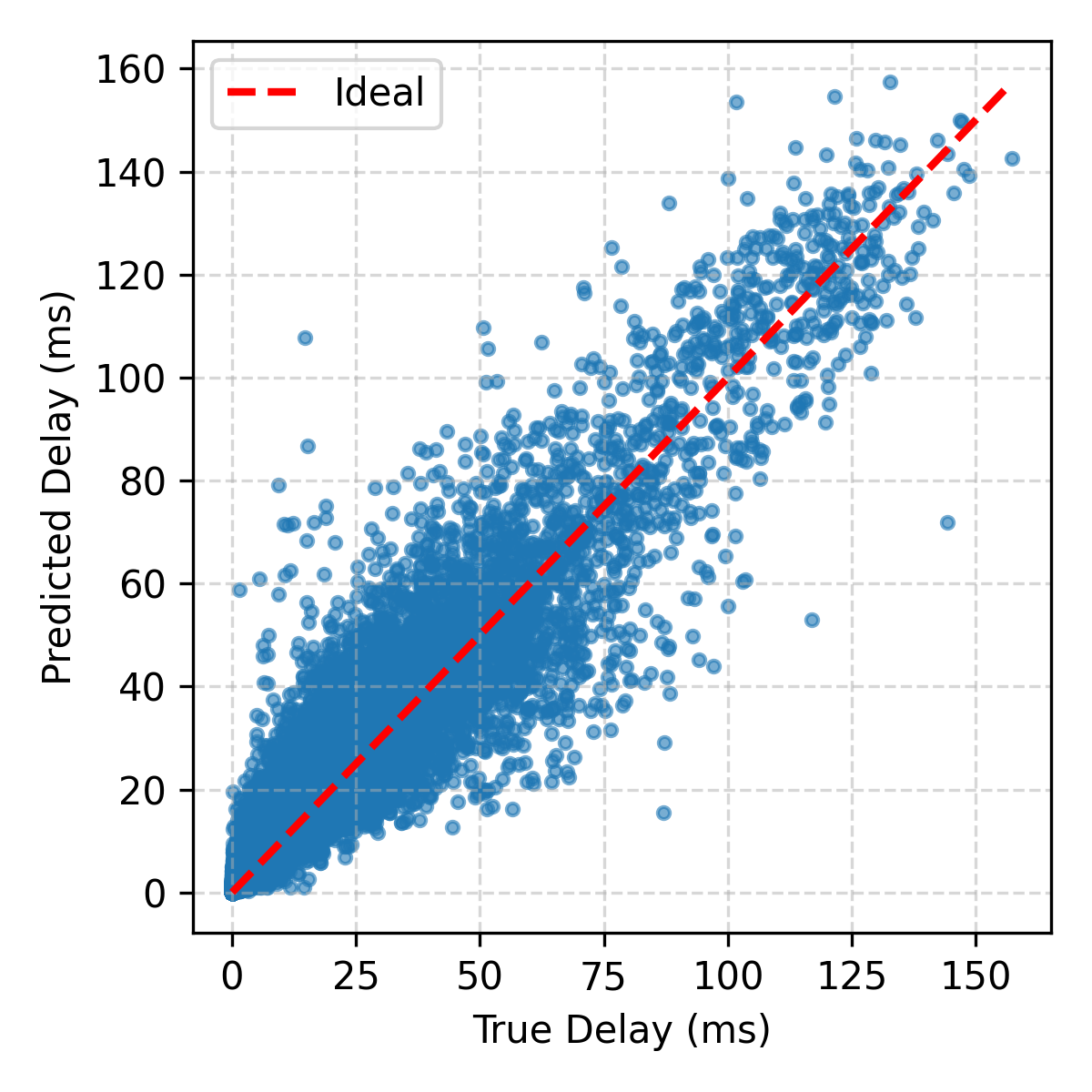}

\caption{Baseline GNN 
}
\end{subfigure}\hfill
\begin{subfigure}{0.30\textwidth}
\centering
\includegraphics[width=\linewidth]{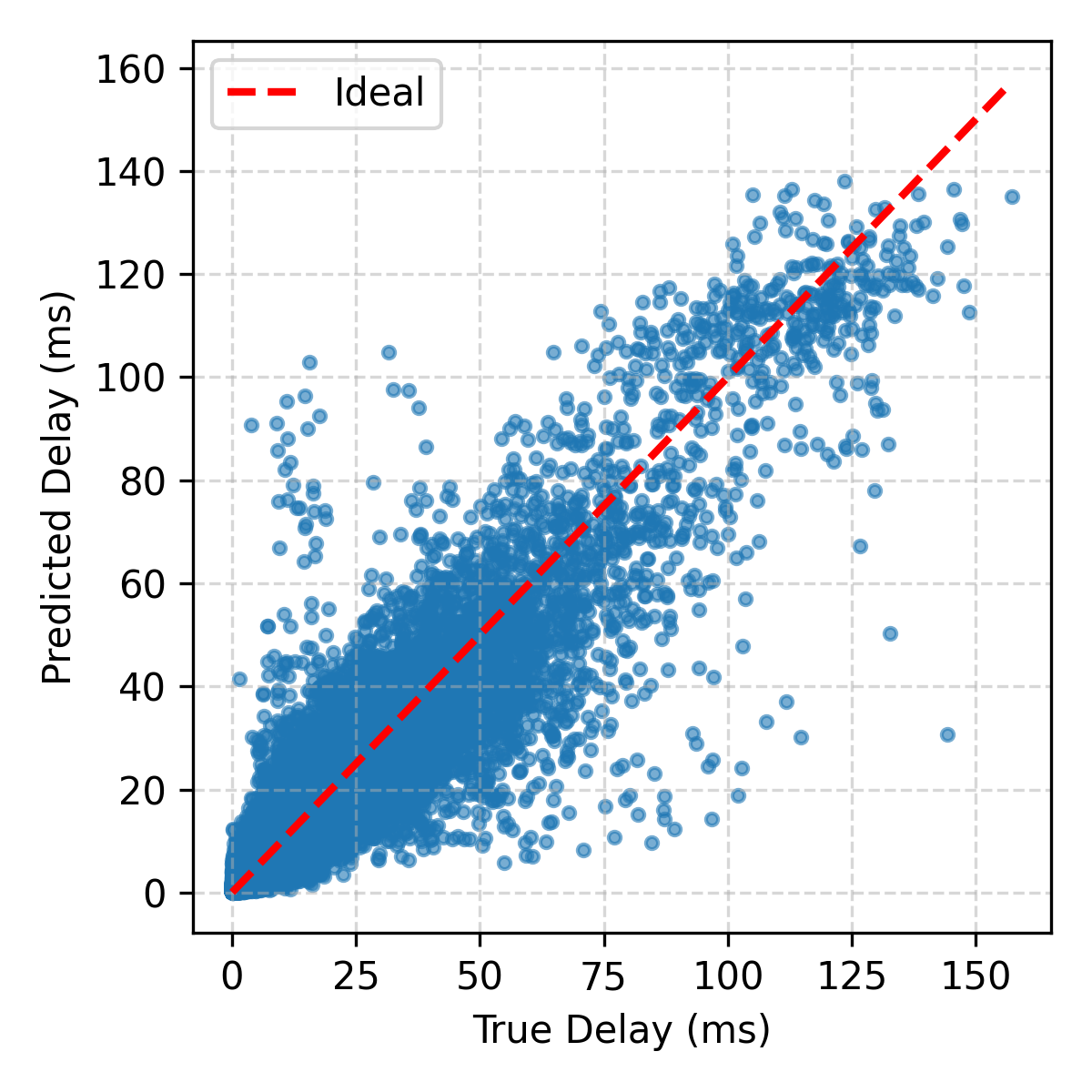}
\caption{FlowKANet 
}
\end{subfigure}\hfill
\begin{subfigure}{0.30\textwidth}
\centering
\includegraphics[width=\linewidth]{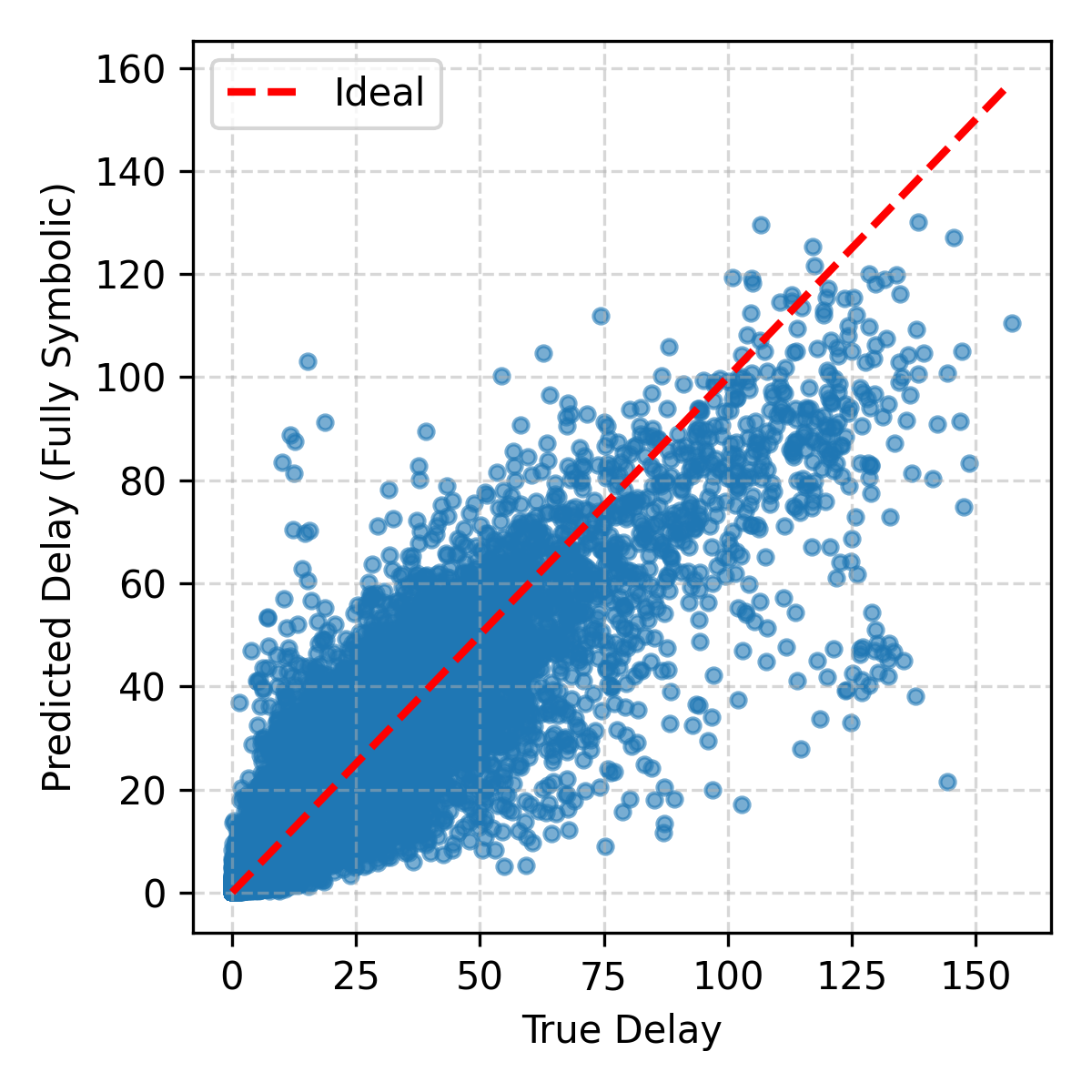}
\caption{Symbolic surrogate 
}
\end{subfigure}
\caption{Predicted vs.\ true per-flow delay on the test set.} 
\label{fig:scatter_all}
\vspace{-2.5em}
\end{figure*}

\begin{figure}[h]
\centering
\includegraphics[width=0.49\textwidth]{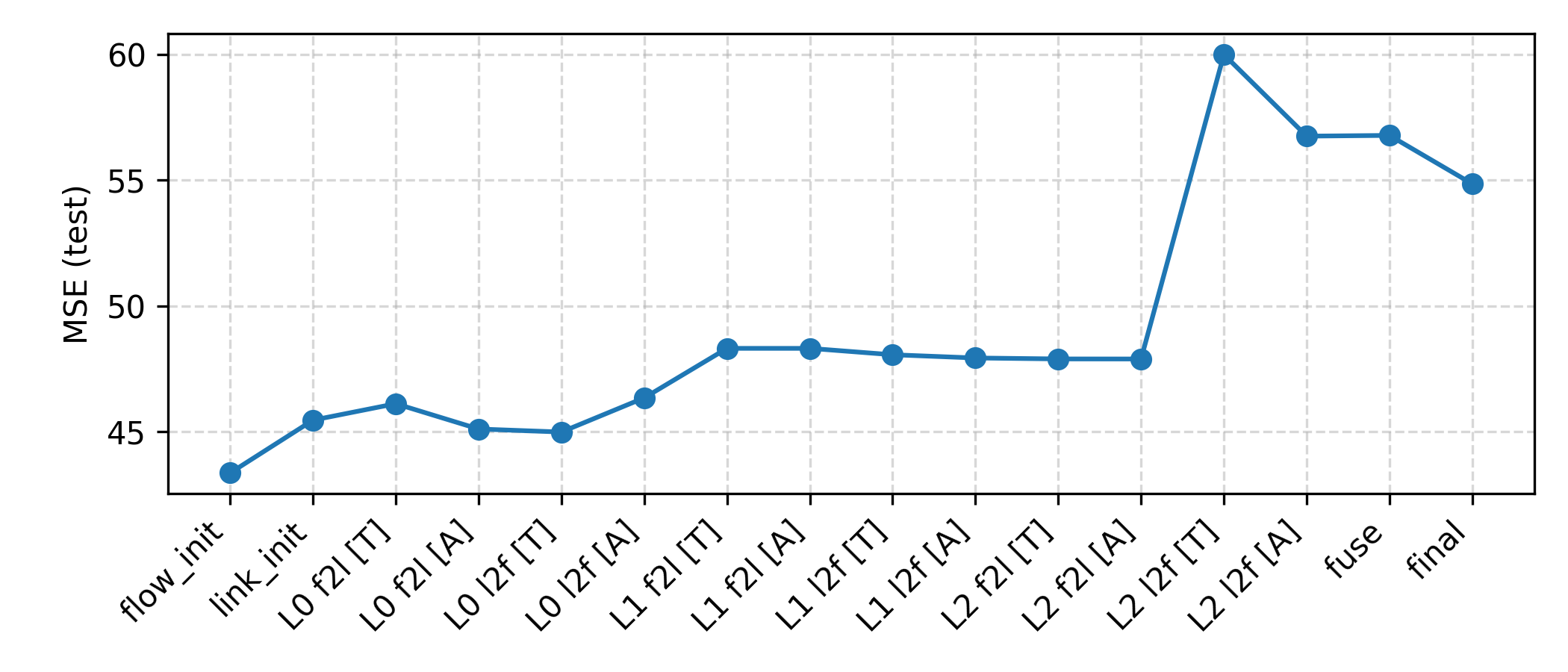}
\caption{Progressive-hybrid full-dataset MSE as blocks are symbolized in sequence (lower is better). \texttt{L0-L2} denote message-passing layers; \texttt{f2l} and \texttt{l2f} refer to flow-to-link and link-to-flow directions.}
\label{fig:progressive_hybrid_mse}
\end{figure}

\begin{table}[h]
\centering
\caption{Test-set predictive accuracy.}
\label{tab:predictive_accuracy}
\begin{tabular}{lcc}
\toprule
Model & MSE (lower is better) & $R^2$ (higher is better) \\
\midrule
Baseline GNN    & 38.6358  & 0.8113 \\
FlowKANet       & 40.8094  & 0.8727 \\
Symbolic surrogate & 54.8562 & 0.8290 \\
\bottomrule
\end{tabular}
\end{table}
Both models achieve strong predictive power on the GNNet dataset.  
FlowKANet attains a higher $R^2$ with slightly higher MSE, showing that spline operators capture small-delay variance and remain stable via the Softplus readout. The symbolic surrogate, though less accurate, provides transparent closed-form equations suited for interpretable or lightweight deployment. Figure~\ref{fig:scatter_all} shows predicted vs.\ true delays; FlowKANet points align closer to the diagonal, confirming improved variance capture. During symbolic distillation, we tracked the model MSE as each KAN block was replaced (“progressive hybrid”). Figure~\ref{fig:progressive_hybrid_mse} shows that replacing early encoders and mid-level message-passing blocks causes mild degradation, while symbolizing the final layers increases error more sharply. Thus, late components are most critical for accuracy, suggesting hybrid deployments, symbolic early/mid blocks with neural readout, offer the best trade-off between interpretability and performance.
Overall, KANs bridge conventional GNNs and symbolic models, reducing parameters while retaining accuracy and enabling transparent surrogates.  
Future work will address accuracy loss in final blocks through symbolic operators for message passing and adaptive hybrid designs.

\subsection{Parameter Efficiency}

Model compactness is critical for deployment. Table~\ref{tab:param_efficiency} compares the number of parameters across the three models. FlowKANet reduces the trainable parameter count by nearly $5\times$ compared to the GNN (20k vs.\ 98k), and the symbolic surrogate eliminates all trainable weights, leaving only 267 fixed constants in the final equations. This progression illustrates a efficiency-interpretability tradeoff: from large but flexible GNNs, through compact FlowKANet, to purely analytical surrogates suitable for constrained or safety-critical environments.

\begin{table}[h]
\centering
\caption{Trainable parameter counts of baseline GNN, FlowKANet, and symbolic surrogate.}
\label{tab:param_efficiency}
\begin{tabular}{l c c}
\toprule
Model & Model Parameters  \\
\midrule
Baseline GNN & 98{,}210 \\
FlowKANet & 20{,}094 \\
Symbolic surrogate & 267 (Constants in Equations) \\
\bottomrule
\end{tabular}
\end{table}
\vspace{-1em}
\section{Conclusion}
We have presented a unified framework for flow delay prediction, covering three levels of model design: a heterogeneous GNN with attention-based message passing, FlowKANet model with spline-based transformations, and fully symbolic surrogates distilled from the KAN model. Our experiments show that the FlowKANet achieves comparable accuracy to the GNN while reducing parameter count nearly five-fold. The symbolic surrogates, although less accurate, eliminate trainable parameters entirely and produce transparent closed-form equations that respect the original graph structure. This progression highlights a clear spectrum of trade-offs: from accuracy-focused neural models, to compact spline-based architectures, to symbolic predictors suitable for deployment in resource-constrained or safety-critical environments. In future work, we will focus on deeper interpretation of the learned transformations, both in the KAN-based model and in the distilled symbolic equations, to provide further insights into how graph-structured dependencies drive flow delay. 
\vspace{-0.6em}

\section*{Acknowledgment}
Supported by the Agence Nationale de la Recherche, France (Grant ANR-21-CE25-0005, SAFE project).
\vspace{-0.2em}
\bibliographystyle{IEEEtran}
\bibliography{biblio}

\end{document}